\documentclass{article}
\usepackage{spconf,amsmath,graphicx}
\usepackage{graphicx}
\usepackage{amsmath}
\usepackage{amssymb}
\usepackage{booktabs}
\usepackage{multirow}
\usepackage{soul}
\usepackage{graphicx}
\usepackage{booktabs}       
\usepackage{epsfig}
\usepackage{amsmath}
\usepackage{amssymb}
\usepackage{algorithm}
\usepackage{xcolor}
\usepackage{wrapfig}
\usepackage{midfloat}
\usepackage{cuted}
\usepackage{hyperref}
\usepackage{algpseudocode}
\algblockdefx{ForAllP}{EndForAllP}[1]%
  {\textbf{for all }#1 \textbf{do in parallel}}%
  {\textbf{end for}} 
\definecolor{commentcolor}{HTML}{93a1a1}

\usepackage{enumitem}
\setlist{nosep, leftmargin=14pt}

\usepackage{mwe} 

\newcommand{\mymethod}{Rate-My-LoRA}

\title{\mymethod: Efficient and Adaptive Federated Model Tuning for Cardiac MRI Segmentation}
\name{\parbox{\linewidth}{\centering{Xiaoxiao He$^{1}$, Haizhou Shi$^{1}$, Ligong Han$^{1*}$, Chaowei Tan$^{1}$, Bo Liu$^{2*}$, Zihao Xu$^{1}$, Meng Ye$^{1}$, Leon Axel$^{3}$, Kang Li$^{1}$, Dimitris Metaxas$^{1}$}}}

\address{$^{1}$Department of Computer Science, Rutgers University, USA\\
$^{2}$Walmart Global Tech, USA\\
$^{3}$New York University School of Medicine, USA
}
\begin{document}
%
\maketitle
\vspace{-0.3cm}
\renewcommand*{\thefootnote}{\fnsymbol{footnote}}
\footnotetext{$^*$ Corresponding Author}
\newcommand{\xu}[1]{\textcolor{blue}{(Xu: #1)}}

\begin{abstract}
Cardiovascular disease (CVD) and cardiac dyssynchrony are major public health problems in the United States. Precise cardiac image segmentation is crucial for extracting quantitative measures that help categorize cardiac dyssynchrony. However, achieving high accuracy often depends on centralizing large datasets from different hospitals, which can be challenging due to privacy concerns. To solve this problem, Federated Learning (FL) is proposed to enable decentralized model training on such data without exchanging sensitive information. However, bandwidth limitations and data heterogeneity remain as significant challenges in conventional FL algorithms. In this paper, we propose a novel efficient and adaptive federate learning method for cardiac segmentation that improves model performance while reducing the bandwidth requirement. Our method leverages the low-rank adaptation (LoRA) to regularize model weight update and reduce communication overhead. We also propose a \mymethod{} aggregation technique to address data heterogeneity among clients. This technique adaptively penalizes the aggregated weights from different clients by comparing the validation accuracy in each client, allowing better generalization performance and fast local adaptation. In-client and cross-client evaluations on public cardiac MR datasets demonstrate the superiority of our method over other LoRA-based federate learning approaches. Code can be accessed \href{https://github.com/hexiaoxiao-cs/Rate-My-LoRA}{here}.
\end{abstract}

%
\section{Introduction}
\vspace{-0.3cm}
\label{sec:intro}
\begin{figure}[t]
    \centering
    \includegraphics[width=1\linewidth]{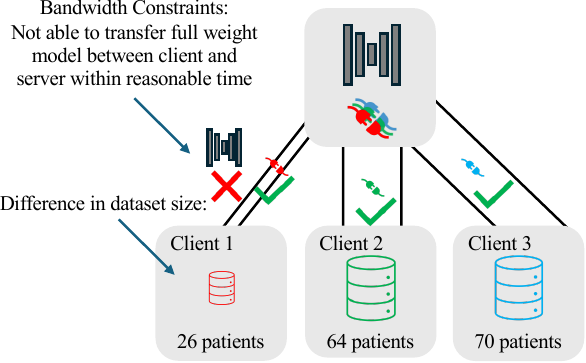}
    \caption{A demonstration of our FL scenario. it shows a scenario where the bandwidth limits the transfer of the model with full weights. Additionally, the dataset exhibits size imbalance; one hospital has half the patients compared to others. } 
    \label{fig:fl}
\end{figure}
\begin{figure}
    \centering
    \includegraphics[width=1\linewidth]{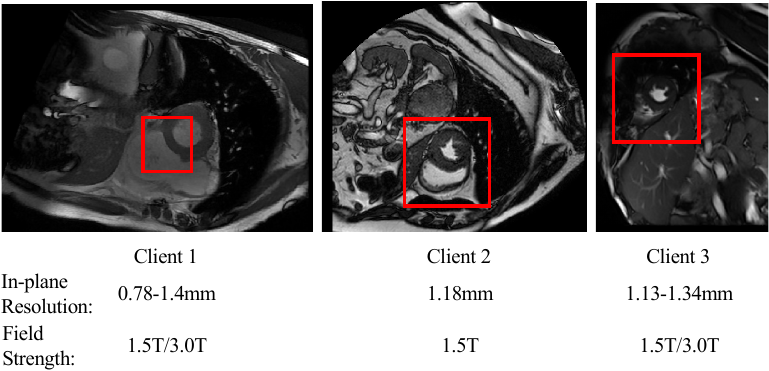}
    \caption{Demonstration of the highly non-IID data on federated learning. Due to different imaging equipment, the style of the images across clients differs, as in the red box.} 
    \label{fig:iid}
\end{figure}

Cardiac dyssynchrony, where the heart's ventricles beat out of sync, worsens heart failure prognosis, with up to 60\% of patients dying within four years \cite{10.1093/ehjci/jes154}. Advances in diagnostic tools like cardiac MRI (cMRI) are crucial in addressing this issue \cite{marcu2006clinical}. Accurate segmentation of the left ventricle cavity (LVC), myocardium (LVM), and right ventricle cavity (RVC) in cMRI is essential for analyzing cardiac dyssynchrony, particularly in measuring LVM wall displacement \cite{he2023dmcvr}. Deep learning has significantly improved automated segmentation \cite{he2019effective,zhangli2022region,liu2022transfusion,he2022recursive}, aiding in this analysis. However, centralized learning methods, which require data sharing across hospitals, raise privacy concerns \cite{chen2020deep}. Federated learning (FL) \cite{FedAVG,pati2022federated,he2023dealing,chang2023mining} addresses these by allowing model training across sites without direct data sharing. While FL protects patient privacy and leverages diverse data, it faces challenges such as data heterogeneity across hospitals~\cite{chen2022fedtune}. Differences in equipment, protocols, and patient populations result in non-IID data, which can hinder model performance. Limited bandwidth in resource-constrained hospitals also complicates participation in FL, potentially excluding valuable data and amplifying biases \cite{sui-etal-2020-feded}, as demonstrated in Fig.~\ref{fig:fl},\ref{fig:iid}.

Researchers have made efforts to address the challenges of highly heterogeneous data, but the issues remain. Stripelis et al.~\cite{stripelis_performance_2022} introduced a federated learning framework where clients' weights are determined by the validation performance of models from other clients. While effective, this method is resource-intensive, requiring each client to validate all other models. Xu et al.~\cite{xu_client_2022} proposed evaluating clients' models on a server using an additional dataset, assigning weights based on performance. However, this may reduce generalization by favoring models that excel on the server's validation set.

\begin{figure*}[t]
    \centering
    \includegraphics[width=0.95\linewidth]{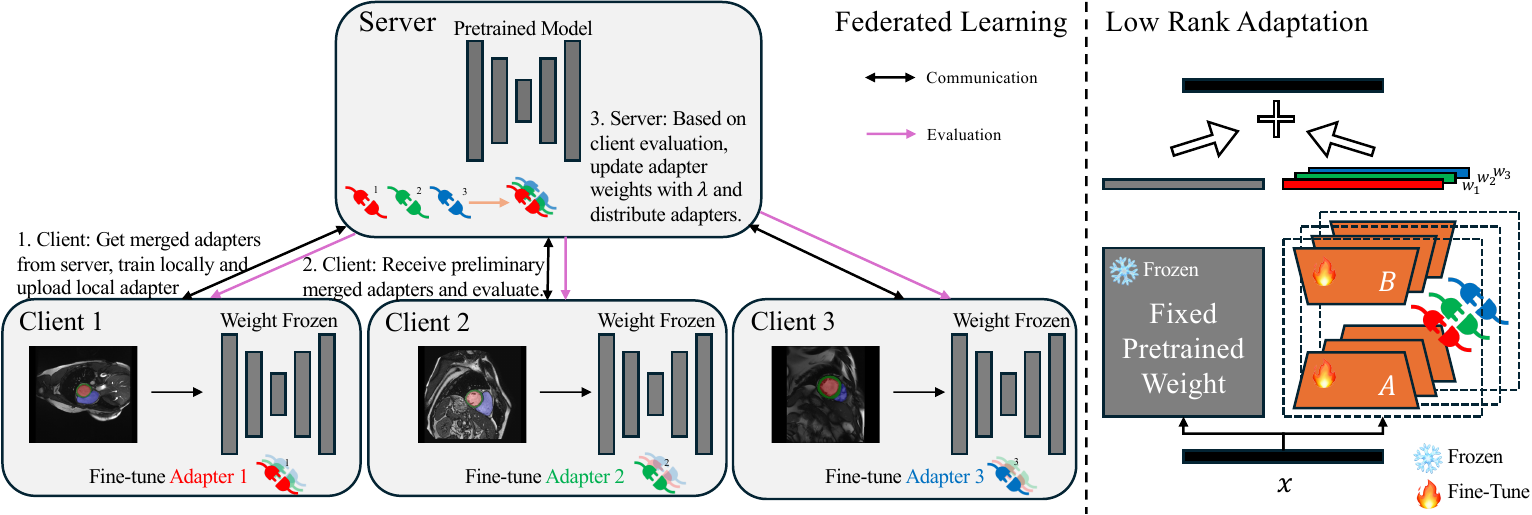}
    \caption{Overview of our efficient and adaptive FL method: the server contains a pretrained model and each client has a local adapter. The red, green and blue label in the MR image represents LVC, LVM, and RVC, respectively. In each FL round, only the adapters in each client are updated. }
    \label{fig:1}
\end{figure*}

Furthermore, previous works that attempted to address the convergence of highly heterogeneous data \cite{stripelis_performance_2022,xu_client_2022} usually overlooked communication efficiency between clients. To tackle this issue, Low-rank adaptation (LoRA) \cite{hu2021lora} has emerged as an effective method for fine-tuning large-scale models in federated learning. Unlike traditional fine-tuning, which updates all model parameters, LoRA introduces trainable low-rank matrices within the model’s weight matrices. This reduces the computational load on clients with limited hardware and significantly cuts communication overhead by transmitting only the compact LoRA parameters instead of full model weights. LoRA’s rank constraints also provide regularization, improving generalization in few-shot scenarios where full-rank fine-tuning may overfit \cite{bau2020rewriting,wang2024blob,zhu2024asymmetry,shi2024training}. FedPETuning \cite{zhang2023fedpetuning} combines LoRA with FedAvg \cite{FedAVG} to achieve similar performance with reduced communication costs, but it fails to account for data heterogeneity, leading to performance loss after merging.


To address this limitation, we introduce \mymethod{}, a novel method designed to enhance the extraction of generalizable features in federated learning and enable efficient local adaptation with minimal training. Our approach begins by evaluating each client's LoRA adapters on their validation sets, followed by applying an adaptive penalization term when merged LoRAs show performance declines. 
This process mirrors the iterative refinement seen in diffusion models, which have proven effective in solving inverse problems and editing \cite{han2024proxedit,he2024dice,stathopoulos2024score}.
This encourages the learning of generalizable knowledge, improving model robustness across diverse data. The method also dynamically adjusts the size of LoRA adapters based on local training set size to prevent overfitting. Our contributions are: (i) \mymethod{} enhances both in-client and cross-client accuracy in LoRA-based FL, (ii) extensive evaluations show superior performance over other methods, and (iii) it reduces bandwidth usage by up to 94\% per communication round.

\section{Methods}
\vspace{-0.3cm}

Fig.~\ref{fig:1} provides a visual representation of our proposed LoRA-based federated learning approach. In essence, we fine-tune a low-rank adapter within each client using local datasets and subsequently merge these adapters using our novel \mymethod{} aggregation method. 

\noindent\textbf{Problem formulation.} We consider a scenario with $|C|$ clients, where each client $c$ possesses a local dataset $D^c$. The goal is to find a global model $\mathcal{M}^*$ that satisfies the following condition: Denote $C$ is the set of all clients, $\mathcal{M}^*=\text{argmin}_{\{\mathcal{M}^c\}}\sum_{c\in C} f(\mathcal{M}^c,\mathcal{D}^c)$. In general, the local objectives measure the local empirical risk over the local dataset $\mathcal{D}^c$ is defined as  $f(\mathcal{M}^c):=\frac{1}{|\mathcal{D}_c|} \sum_{x\in \mathcal{D}^c} f(\mathcal{M}^c;x)$, $\mathcal{M}^c=\mathcal{M}_0+\sum_{k\in C, k\neq c}\mathcal{A}^k+\mathcal{A}_{T+1}^c$ where $\mathcal{M}_0$ is the pretrained model, $T$ is the communication rounds, $\mathcal{A}$ is the learned low-rank adapter.

\noindent\textbf{Federated learning framework with LoRA aggregation.} In order to allow more hospitals to participate in training and remove the barrier of communication constraint, we integrate the federated learning framework with Low-Rank Adaptation. With federated learning, no personally-identifiable patient data will be transferred in any part of the training process. The core idea of LoRA is to constrain the weight update on the model by a low rank decomposition: $W=W_0+\mathcal{A} = W_0+BA$, where $W_0\in \mathbb{R}^{d\times k}$ and $B\in \mathbb{R}^{d\times r}, A\in \mathbb{R}^{r\times k}$ and $r\ll\min({d,k})$. By using low-rank decomposition, the number of parameters requiring fine-tuning and transmission between client and server is drastically reduced compared to full-weight fine-tuning. This not only alleviates communication bottlenecks but also acts as a form of regularization, limiting the model's capacity to memorize local datasets. Consequently, the generalizability of the aggregated model is enhanced, particularly in few-shot scenarios where overfitting is a concern~\cite{bau2020rewriting,zhu2024asymmetry,han2023svdiff,zhang2024spectrum}, as utlined in Alg.~\ref{alg:fedavg}.
\begin{figure*}[t]
    \centering
    \includegraphics[width=0.75\linewidth]{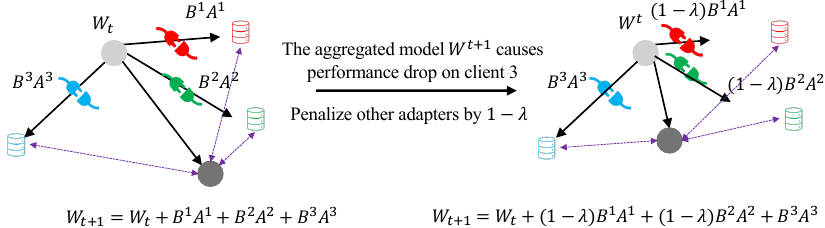}
    \caption{Our \mymethod{} method that utilizes on-client data for evaluating the performance gains/losses of the aggregated model and adaptive adjust the aggregation weight of each adapter to achieve overall performance. }
    \label{fig:2}
\end{figure*}
\begin{algorithm}
	\caption{The proposed \mymethod{} method}
	\label{alg:fedavg} 
	\begin{algorithmic}[1] 
	    \item[\textbf{Central server do:}]
	        \State Load pretrained network $\mathcal{M}$.
            \State Inject and initialize LoRA adapters $\forall c\in C, B^c_0,A^c_0 $ to $\mathcal{M}$.
            \State Distribute $\mathcal{M}, B^c_0,A^c_0$ to clients.
	        \For {each communication round $t \in {1, ..., T}$}
	            \ForAllP {each client $c \in C$}
	                \State $B_{t}^{c}A_{t}^{c} \leftarrow$ TrainLocal$(\mathcal{M}+\sum_{k\in C}B^k_{t-1}A^k_{t-1})$ \Comment{\textcolor{commentcolor}{Collect adapter from client $c$}}
	            \EndForAllP
                \ForAllP {each client $c\in C$}
                    \State $P_c^t \leftarrow$ Evaluate $\mathcal{M}_t^A$ on client $c$ and return the evaluation metric
                \EndForAllP
                \State Update $B_{t+1}^c$ with \mymethod{} adaptive learning weight $w_t(c)$, indicated in Eqn.~\ref{eqn:wtc}
                \State Distribute $B_{t+1}^c,A_{t+1}^c$
                \State $\lambda=\lambda \times0.95$ \Comment{\textcolor{commentcolor}{Update lambda}}
	        \EndFor 
	\end{algorithmic} 

\end{algorithm}

\noindent\textbf{The proposed \mymethod{} method.} With conventional methods for merging LoRA adapters, includes average weighting $\mathcal{M}=\mathcal{M}_0+\frac{1}{|C|}\sum_{c\in C}\mathcal{A}^c$ or weighting by dataset size $\mathcal{M}=\mathcal{M}_0+\frac{1}{\sum_{c\in C}|\mathcal{D}^c|}\sum_{c\in C}|\mathcal{D}^c|\mathcal{A}^c$ (FedPETuning), it is common to see a drastic accuracy drop after merging the adapters. In average weighting, the aggregated model often becomes biased towards clients with shared features, as their models converge in similar directions during training. This can lead to an overemphasis on common features, reducing the model’s ability to generalize to unique client data. With FedPETuning, where merging weights are based on dataset size, the model risks overfitting to the client with the most data. This dominant client’s model can overshadow others, resulting in a model that performs well on their data but poorly on others, limiting overall generalization.



Both conventional methods compromise the generalizability of the collaborative model in different ways. To overcome this and improve the robustness of the aggregated model, we propose a novel approach. After client-side training, each client uploads its learned LoRA adapter weights to the server, which then redistributes them to all clients. Each client evaluates the aggregated model on its local validation set and reports the accuracy back to the server. This accuracy score indicates how well each adapter captures common features across clients. By comparing the current accuracy to the previous round, we can penalize adapters that reduce generalizability by introducing a penalty term. The weight for each LoRA adapter is determined using the hyperparameter $\lambda$, defined as follows:
\begin{gather}
    w_t(c)=\begin{cases}
        1-\lambda & \exists i \in \mathcal{C} ~\text{s.t.}~ P^i_t<P^i_{t-1} ~\text{and}~ P^c_t>P^c_{t-1}\\ 
        1 & \text{otherwise}
    \end{cases}
    \label{eqn:wtc}
\end{gather}
where $P^c_t$ is the client $c$ validation accuracy at time $t$. Note that $w_t(c)=1-\lambda$ only happens when there exists a client that has drop in accuracy. Then for each of the weight matrix $B^c_tA^c_t$, the aggregation becomes 
\begin{gather}
    W_t = W_{t-1}+\frac{1}{\sum_{c\in \mathcal{C}}|\mathcal{D}^c|}\sum_{c\in \mathcal{C}}{w_t(c)|\mathcal{D}^c|B^c_tA^c_t}
\end{gather}

During the FL aggregation phase, the adapters are first aggregated with equal weights and distributed to each client, which then evaluates the model on its local validation set and reports accuracy to the server. The server tracks validation performance over time. Sometimes, due to data heterogeneity between clients, validation accuracy decreases after adapter aggregation, as shown on the left in Fig.~\ref{fig:2}. The arrows indicate gradient directions post-local training, and the drop in performance is represented by the distance between $W_t$ and $W_{t+1}$ from the blue local optimal point, highlighting conflicts in client gradient descent directions. When such performance drop is reported, the server adaptively penalizes the adapter from other clients, as they fail to generalize. As illustrated on the right of Fig.~\ref{fig:2}, this reduces the distance between the aggregated model and local optima, improving overall model performance. To ensure convergence, we apply a diminishing schedule to $\lambda$, reducing it by 5\% each communication round. When $\lambda=0$, the method reverts to FedAVG, where LoRA adapters are averaged across all clients.

\begin{table}[t]
    \centering
    \begin{tabular}{cccccc}
\toprule
Method &$\mathcal{C}$& DICE & VOE & HD & ASSD\\ 
 \hline
\multirow{3}{*}{Local Only} &1 & 0.818 & 29.957 & 89.897 & 3.769\\ 
 & 2& 0.830 & 27.503 & 93.124 & 4.171\\ 
 &3& 0.887 & 19.984 & 56.680 & 1.047\\ 
\hline
 \multirow{3}{*}{\parbox{2cm}{\centering Average Weighting$^\dagger$ \cite{FedAVG}}} &1& 0.834 & 26.990 & 12.652 & 0.994\\ 
 &2& 0.896 & 18.650 & 54.945 & 1.111\\ 
 &3& 0.834 & 28.110 & 74.786 & 2.884\\
 \hline
\multirow{3}{*}{\parbox{2cm}{\centering FedPETuning \cite{zhang2023fedpetuning}}} &1& 0.840 & 27.339 & \textbf{8.402} & 0.815\\ 
 & 2& 0.883 & 19.947 & \textbf{41.096} & 2.040\\ 
 &3& 0.873 & 21.913 & 36.701 & 0.943\\ 
 \hline 
\multirow{3}{*}{Our method} &1& \textbf{0.889} & \textbf{19.777} & 14.992 & \textbf{0.544}\\ 
 & 2&\textbf{0.910} & \textbf{16.114} & 44.217 & \textbf{1.304}\\ 
 & 3&\textbf{0.895} & \textbf{18.596} &\textbf{26.125} & \textbf{0.688}\\ 
 \hline 
\parbox{2cm}{\centering Full Dataset (Average)} &all& 0.904 & 17.068 & 37.983 & 0.830\\
\bottomrule
    \end{tabular}
    \caption{Quantitative results for in-client evaluation. $^\dagger$: full weights fine-tuning. Full Dataset utilizes all available data to train and evaluate on testing set for all three clients, providing an upper bound. The rest results are evaluated on the client-side testing set. Local only fine-tunes the model with local data without federated learning. The best results have been highlighted in the chart.} 
    \label{tab:personalization}
\end{table}
\begin{table}[t]
    \centering
    \begin{tabular}{cccccc}
\toprule
Model & $\mathcal{C}$ & DICE & VOE & HD & ASSD\\ 
 \hline
\multirow{3}{*}{Local Only} & 1 & 0.801 & 30.505 & 74.569 & 2.716\\
 & 2 & 0.848 & 25.386 & 49.832 & 1.579\\
 & 3 & 0.776 & 33.231 & 81.534 & 3.951\\
\hline
\multirow{3}{*}{\parbox{2cm}{\centering Average Weighting$^\dagger$ \cite{FedAVG}}} & 1 & 0.768 & 35.695 & 17.628 & 1.260\\ 
 & 2 & 0.843 & 26.576 & 27.330 & 1.126\\ 
 & 3 & 0.813 & 30.690 & 90.586 & 2.913\\
 \hline
\multirow{3}{*}{\parbox{2cm}{\centering FedPETuning \cite{zhang2023fedpetuning}}} & 1 & 0.849 & 24.917 & \textbf{29.130} & \textbf{1.018}\\
 & 2 & 0.865 & 23.432 & 20.331 & 0.747\\
 & 3 & 0.862 & 23.429 & 39.445 & 1.211\\
\hline
\multirow{3}{*}{Our Method} & 1 & \textbf{0.882} & \textbf{20.411} & 37.484 & 1.396\\
 & 2 & \textbf{0.894} & \textbf{18.826} & \textbf{13.139} & \textbf{0.557}\\
 & 3 & \textbf{0.904} & \textbf{17.187} & \textbf{24.317} & \textbf{0.697}\\
\bottomrule
    \end{tabular}
    \caption{Quantitative results for cross-client accuracy. The number in $\mathcal{C}$ column indicates that the model is trained on such client and is evaluated on the other two clients. The best results have been highlighted in the chart.}
    \label{tab:generalization}
\end{table}

\begin{figure}
    \centering
    \includegraphics[width=\linewidth]{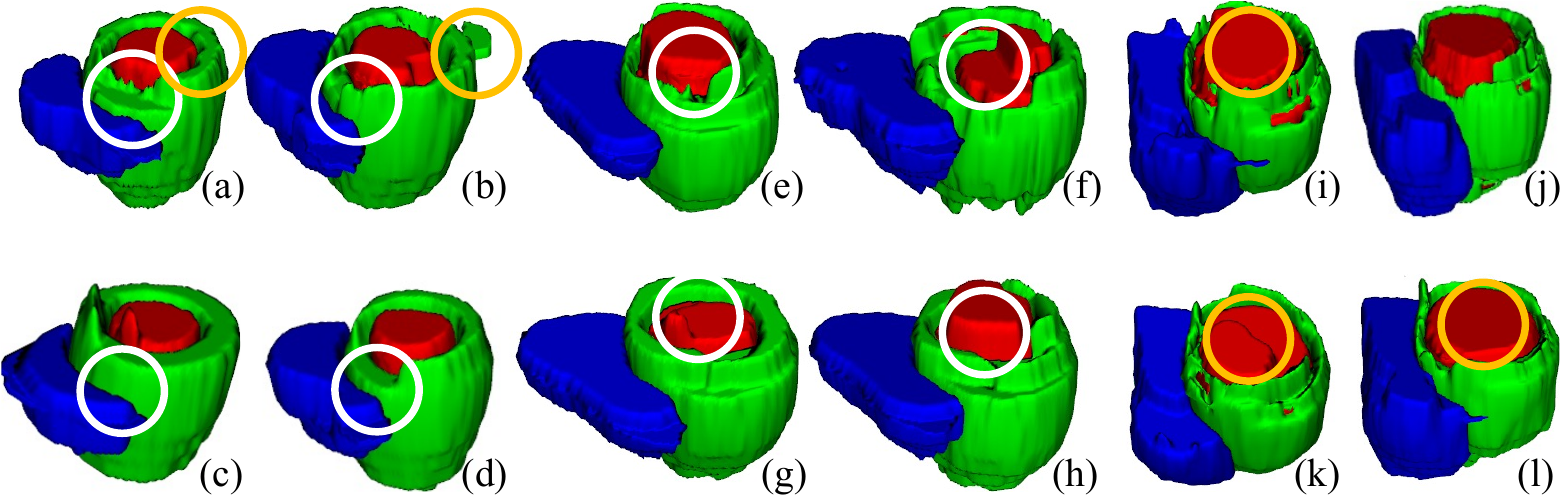}
    \caption{3D visualization results of in-client evaluation. (a-d), (e-h), (i-l) are the ground truth, local training, FedPETuning and the \mymethod{} method results from the client 1, 2 and 3, respectively.}
    \label{fig:3d}
\end{figure}

\begin{figure}[t]
    \centering
    \includegraphics[width=\linewidth]{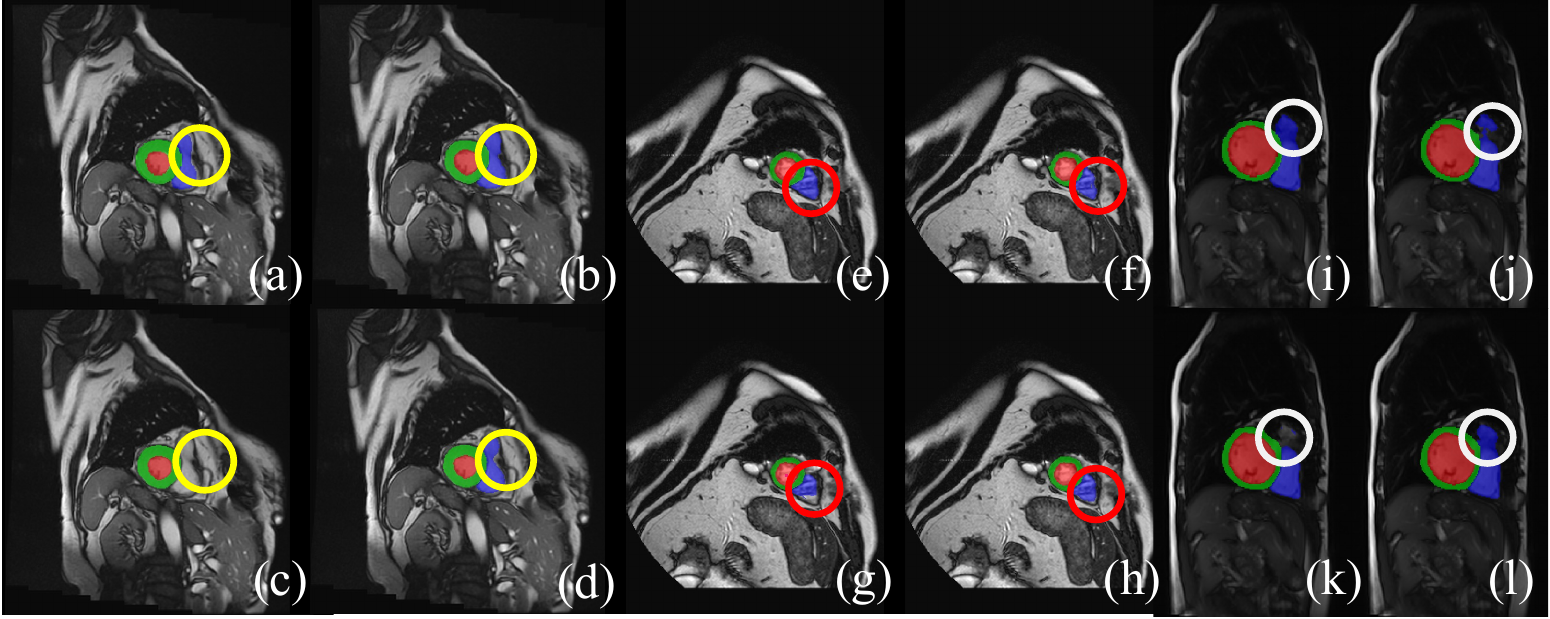}
    \caption{Visualization of the in-client evaluation results. (a-d), (e-h), (i-l) are the ground truth, local training, FedPETuning and our method results from client 1,2 and 3,}
    \label{fig:visualization-result-1}
\end{figure} 
\section{Experiments}
\vspace{-0.3cm}

\textbf{Experimental Settings.}
This paper uses two publicly available datasets: the Automated Cardiac Diagnosis Challenge (ACDC) \cite{bernard2018deep} and the Multi-Disease, Multi-View \& Multi-Center Right Ventricular Segmentation in Cardiac MRI (M\&Ms-2) \cite{campello2021multi,martin2023deep}. The ACDC dataset, with 100 patients imaged using a Siemens scanner, is used to train the base model. The M\&Ms-2 dataset includes cardiac cine MR images from 160 patients across three hospitals using GE (Client 1), Philips (Client 2), and Siemens (Client 3) scanners, creating a visual appearance gap between FL clients. All images are cropped to $224\times 224$, and intensity is normalized. A train/validation/test split of 8:1:1 is applied. U-net \cite{ronneberger2015u} serves as the base model, with LoRA adapters injected into each convolution block. LoRA adapter sizes are $16,32,32$, based on local dataset size, and $\lambda$ is set to $0.2$. After aggregation, models are fine-tuned on local data for one extra epoch. We evaluate performance using Dice coefficient (DICE), volumetric error (VOE), Hausdorff distance (HD), and average symmetric surface distance (ASSD) between ground truth and segmentation results.

\noindent\textbf{Evaluation of the segmentation quality.}
In Tab.~\ref{tab:personalization}, we present the in-client evaluation, where our method outperforms FedPETuning by up to 4.7\% on client 1. The lower performance of FedAVG with full weights, compared to local fine-tuning on client 3, highlights how data heterogeneity causes performance drops in the aggregated model. Across the full-weight and LoRA-based FL methods, we observe a general performance improvement and reduced client inconsistency, suggesting that low-rank regularization enhances the model’s generalizability. Our \mymethod{} approach helps each client extract generalizable knowledge, boosting segmentation accuracy and narrowing the performance gap between clients. In Fig.~\ref{fig:visualization-result-1}, we show the in-client evaluation results visually. FedPETuning and local training struggle to segment the RVC. Our method successfully segments the RVC, LVC, and LVM regions. Additionally, Fig.~\ref{fig:3d} shows 3D visualizations, where other methods exhibit varying degrees of oversegmentation or undersegmentation, while our \mymethod{} produces volumes closest to the ground truth.

In Tab.~\ref{tab:generalization}, we present cross-client accuracy results. Consistent with Chen et al. \cite{chen2022fedtune}, non-IID conditions between clients degrade the performance of Full-weight FedAVG, compared to local training. Using low-rank regularization and our adaptive \mymethod{}, we show that knowledge extracted from one client improves other clients’ models, enhancing the generalizability of the aggregated model. We want to emphasize that in terms of bandwidth usage, our method saves up to 15.5x compared to full-weight training per epoch, as calculated by the size of LoRA adapters ($1.8$MB, $3.6$MB, $3.6$MB on each client respectively) and the full rank weight size (28MB). Although it requires multiple communications, LoRA weights are only uploaded and downloaded once per iteration.

\section{Conclusion and Discussion}
In this paper, we proposed a FL-based low-rank adaptation method, \mymethod{}, to improve both in-client and cross-client cardiac segmentation accuracy under communication bandwidth constraints. \mymethod{} carefully evaluates the generalizability of LoRA adapters from different clients and enables fast local adaptation. Both in-client and cross-client evaluations show that our method outperforms other LoRA-based FL approaches. This work holds potential for enabling medical institutions in resource-limited settings to train AI models effectively. 

\section{Compliance with ethical standards}
\label{sec:ethics}

This research study was conducted retrospectively using human subject data made available in open access by Universitat de Barcelona, Spain and University of Lyon, France. Ethical approval was not required as confirmed by the license attached with the open access data.

\section{Acknowledgments}
\label{sec:acknowledgments}
This research has been partially funded by research grants to D. Metaxas through NSF: IUCRC CARTA 1747778, 2235405, 2212301, 1951890, 2003874, and NIH-5R01HL127661.

\bibliographystyle{IEEEbib}
\bibliography{strings,refs}

\end{document}